\documentclass[journal]{IEEEtai}

\usepackage{algorithm}
\usepackage{algpseudocode}
\usepackage{multirow}
\makeatletter
\global\let\ModernLaTeXBegin\begin
\global\let\ModernLaTeXEnd\end
\global\let\ModernLaTeXDocument\document
\global\let\ModernLaTeXEndDocument\enddocument
\makeatother
\usepackage{arabtex}
\usepackage{csvsimple}
\usepackage{svg}
\usepackage{amsmath}
\usepackage{hyperref} 
\usepackage{caption}

\usepackage{utf8}
\setcode{utf8}
\makeatletter
\global\let\begin\ModernLaTeXBegin
\global\let\end\ModernLaTeXEnd
\global\let\document\ModernLaTeXDocument
\global\let\enddocument\ModernLaTeXEndDocument
\AtBeginDocument{\setarab\setnone\input{apatch.sty}}
\makeatother

\title{EXTRACTIVE SUMMARIZATION FOR ARABIC DOCUMENTS USING SARABERT WITH A SEMANTIC SIAMESE SIMILARITY EVALUATION METRIC}

\author{ Sami Shames El Deen
\and , Mariette Awad\\
\small American University of Beirut}
\date{}

\begin{document}
\maketitle




\begin{abstract}
In this research, we introduce SAraBERT, an enhanced version of AraBERT which proposes inter-sentence transformer layers for extractive summarization tasks. To ensure that the summaries generated by SAraBERT achieve a high coverage of the document's main ideas, we propose  Semantic Siamese Similarity, a novel evaluation metric that measures the level of similarity between two text inputs. We validated using BLEU, ROUGE, and Semantic Siamese similarity on Sarabert and published related models. Simulation results showed the effectiveness of our proposed model and motivate follow on research.
\end{abstract}

\begin{IEEEImpStatement}
This paper introduces SAraBERT, an improvement on Arabert, that incorporates inter-sentence transformer layers to perform extractive summarization on Arabic text. Furthermore, a novel called Semantic Siamese Similarity (SSS) is proposed to combine embedding similarity and exact match. Experiments using various evaluation metrics show that SAraBert outmatches its predecessors on Arabic extractive summarizing reaching new state of the art levels of performance.
\end{IEEEImpStatement}





\section{Introduction}\label{chap:introduction}
\IEEEPARstart{A}{s} the amount of textual information available online grows rapidly, it becomes difficult for readers to read large amounts of text and find out which of these texts are useful. As a result, researchers in the field of automatic text summarization need tools that can make multiple-document reading more efficient. The task of text summarization is considered one of the most important and challenging NLP tasks. Summarization generates short text from long text, so the short text contains the most important information from the original text. The task is often divided into two paradigms known as extractive summarization and abstractive summarization. The first methodology determines the essential sections of the text using statistical tool. The summary is represented by truncating and connecting these sections. The second methodology emulates human activity in summarizing, which is based on presenting the text’s core idea in a new linguistic style and using different terms. It incorporates more complicated procedures including paraphrase, generalization, and reordering \cite{jing2002using}.This work focuses on extractive summarization. While there is a wealth of research in the area of extractive summarization, most of this research is based on English texts, and there is a lack of research on summarizing Arabic texts. 
\par Arabic natural language processing (NLP) \cite{Hajj2013}\ is considered more complex than English and other European languages. The main reason for this complexity is the highly derived and rich form of Arabic morphology. This creates a set of challenges, which include: \par 1) Morphological richness \cite{al2016automatic}\cite{al2015multi}:  Arabic is heavily derived, inflected and has a significant impact on NLP tasks such as stemming and lemming. \par 2) Absence of diacritics: diacritics play an important role in determining the meaning of words and facilitating the task of tokenizing and parsing text. \par 3) No capitalization in Arabic language: without the usage of uppercase letters in Arabic, it will be difficult to identify proper nouns, titles, and abbreviations.

\par In this work, we introduce SAraBERT, a novel and enhanced version of AraBERT that adds inter-sentence transformer layers for extractive summarization tasks. To 
ensure that the summaries generated achieve a high coverage of the document’s
main ideas, we also propose a novel evaluation metric, Semantic Siamese Similarity that measures the level of similarity between two text inputs using contextualized embeddings in addition to exact match. We use an English benchmark dataset (CNN/Daily Mail) and translate it to Modern Standard Arabic language to train SAraBERT. Testing the performance of Sarabert in comparison with other published models was evaluated using BLEU, ROUGE, and Semantic Siamese similarity on Kalimat Dataset. Results have shown the effectiveness of our proposed model. \par In summary, the main contributions of this work are as follows:

\begin{itemize}
    \item A novel Arabic language model for extractive summarization that builds on AraBERT model.
    \item A novel hybrid similarity metric that measures similarities between 2 text based on the semantic and syntactic features.
    \item An arabic corpus translated from English that targets the extractive summarization and Question Answering tasks, available upon request.
\end{itemize}

The structure of the paper is such that, in section 2, we survey the related work and background information required for this work. Section 3 describes the model's architecture and the proposed metric. Section 4 shows the experimental research and highlights the obtained results along with some insights. Finally Section 5 concludes the paper with follow on research directions. 
\section{Literature Review}\label{chap:literature}
To provide the necessary context to the proposed solution, it is worth investigating previous research.
\subsection{Classical Approaches}
Until 2017, statistical methods were predominantly used in text summarization which are based on the concept of relevance score and Bayesian classifiers \cite{patil2004statistical}.
Word Frequency approach is the most used methodology for sentence scoring where the sentence score is calculated by the sum of the frequencies while avoiding all stop words. The proposed solution in \cite{alotaiby2012new} generates news titles by combining Word-Frequency, sentence position and similarity measures methodologies. 
Term Frequency-Inverse Document Frequency (TF-IDF) is a numerical methodology that represents the importance of a word to document in a collection of documents (corps) \cite{ferreira2013assessing}. TF-IDF is an improvement on the Word Frequency and shows how the weights are distributed on the words of a document. TF-IDF is used frequently in auto-summarization systems \cite{al2009arabic} \cite{haboush2012arabic} \cite{el2014multi} \cite{fejer2014automatic}.
An Arabic summarization system called AATSS \cite{hewahi2012automatic} adopted an extractive text summarization approach that was mainly based on sentence weighting and scoring. However, it highly depends on the size of the document in terms of number of sentences, which leads to generation of improper grammar context, and it lacks automatic Arabic entity recognition system and Arabic pronoun resolution system.
The classical statistical approaches are used for both single and multi-document summarization and can be used to enhance the selection of important sentences or the elimination of redundant sentences but fail to understand the text, since they only depend on statistical measures \cite{el2011multi}. \cite{miller2019leveraging} proposed a solution for English language by tokenizing the text into clean sentences, then passing the sentences into a BERT model to generate embeddings. Then using K-means they generated clusters and selected summary sentences based on the closest sentence embedding to each cluster centroid.

\subsection{Machine Learning Approach}
Text summarization is considered as a binary classification problem, where a set of documents and their extractive summaries are used as a training set, and each sentence is classified as a summary sentence or non-summary based on statistical, semantic features or a combination of them  \cite{fattah2009ga} \cite{belkebir2015supervised} \cite{al2018hybrid} \cite{nenkova2012survey}. Machine learning methods can be suitable for document summarization as several methods have been introduced and proven to be effective in performing summarization tasks. We will discuss sequence to sequence models and encoder decoder architectures in what follows. 
\\
\textbf{Sequence-to-Sequence Model}
\\
The neural abstractive summarization with sequence-to-sequence models  eme\-rged in \cite{bahdanau2014neural} \cite{sutskever2014sequence}. This approach has been applied to tasks such as headline generation \cite{rush2015neural} and article summarization \cite{nallapati2016classify}. Chopra et al. \cite{chopra2016abstractive} show that attention approaches that are more specific to summarization can further improve the performance of models. Molham et al. \cite{al2020arabic} re-implemented the sequence-to-sequence framework on the Arabic language, which had not witnessed the employment of this model in the text summarization before. However, the authors state that the work still requires expanding the dataset to cover more articles, and to train new models that are beneficial with the Arabic language since it has a unique grammatical language written from right to left.
\\
\textbf{Encoder-Decoder Model}
\\
Google AI's pre-trained language model called Bidirectional Encoder Representations from Transformers (BERT), proved highly efficient at language understanding and achieved convincing results in most NLP tasks. The Arabic language model ARABERT based on BERT for Arabic language was evaluated on Named Entity Recognition, Sentiment Analysis and Question Answering \cite{antoun2020arabert}. Abdulla et al. \cite{abu2020arabic} proposed an extractive Arabic text summarizer based on ARABERT to summarize the Arabic documents by evaluating and extracting the most important sentences at a document. This proposed approach generated a good summary by extracting the most important sentences from paragraphs. However, the model highly depends on sentence boundaries, its coverage accuracy decreases when the text is too long, and extracted sentences sometimes contain linguistic expressions that creates ambiguous summaries.
\subsection{Other Approaches}
Liu et al. of \cite{liu2019fine} managed to solve the problem of sentence boundary detection by adding an additional layer in the embedding that specifies the beginning and end of each consecutive sentence inserted into the encoder.

Go et al. \cite{inoue2021interplay} explored the effects of language variants, data sizes, and fine-tuning task types in Arabic pre-trained language models. The results suggested that the variant proximity of pre-training data to fine-tuning data is more important than the pre-training data size. Moreover, other design decisions can be explored that may contribute to the fine-tuning performance, including vocabulary size, tokenization techniques, and additional data mixtures.

Nasser et al. \cite{zalmout2017don} presented an LSTM-based morphological disambiguation system for Arabic which has significantly outperformed the state-of-the-art systems. The paper suggested exploring additional deep learning architectures for morphological modeling and disambiguation, especially  sequence-to-sequence models. It also suggested to further investigate the role of syntax features in morphological disambiguation, and explore additional techniques for more accurate tagging.

Reda et al. \cite{Reda2022} developed a hybrid transformer based approach for Arabic text summarisation.  First, they performed extractive summarization using a AraBert based sentence embedding approach to determine which sentences of an article are most suited to be part of the summary. Then, they performed abstractive summarization on the extracted summary by feeding it into the MT5 Arabic transformer. They tested their model on the ESAC dataset and got a precision of 53\%, a recall of 55\%, and an f1 score of 49\%. The main shortcoming of this work are as follows. First, pretrained models were used meaning that the models were not fine tuned specifically to perform summarization. Second, the  ESAC dataset is limited to only 153 articles, and 700 summarizations. Using specialized models and a larger dataset may further improve the results.

\section{Proposed Methodology}\label{Methodology}
\subsection{SAraBERT}
\begin{figure}[H]
    \centering
    \includegraphics[width=\columnwidth]{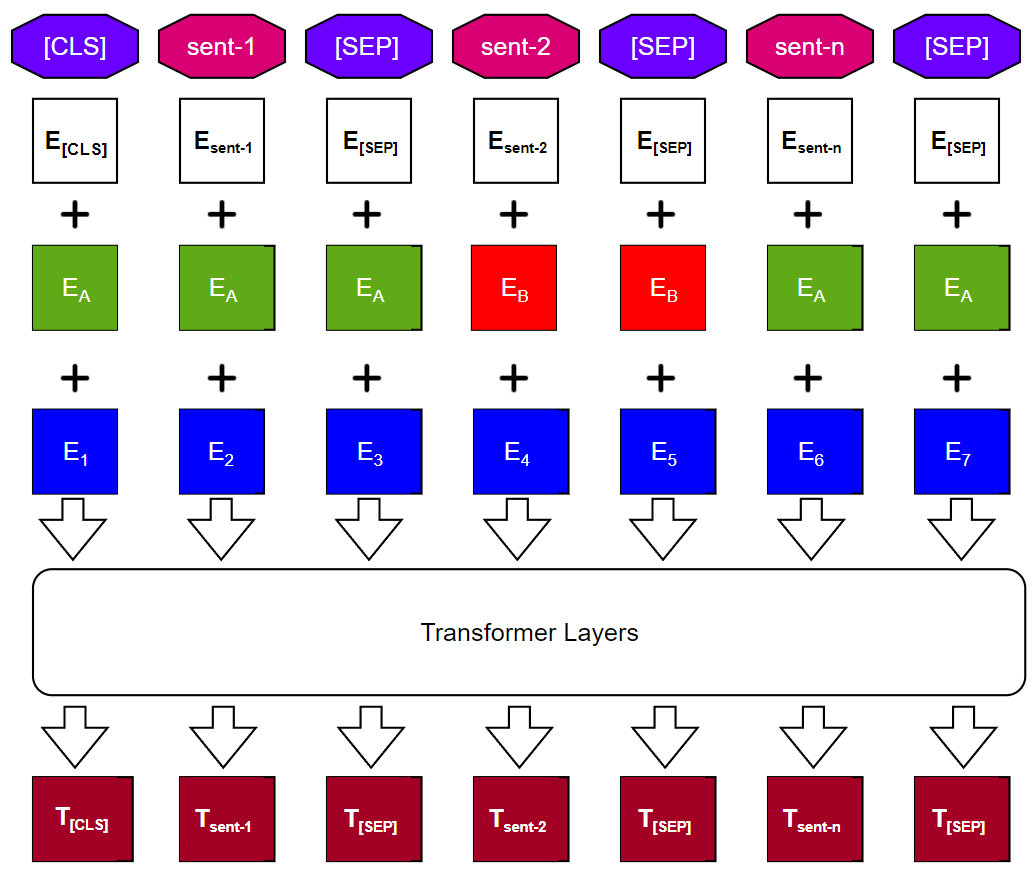}
    \caption{Overview for the architecture of the original BERT model}
    \label{fig:bert}
\end{figure}

\begin{figure}[ht!]
    \centering
    \includegraphics[width=\columnwidth]{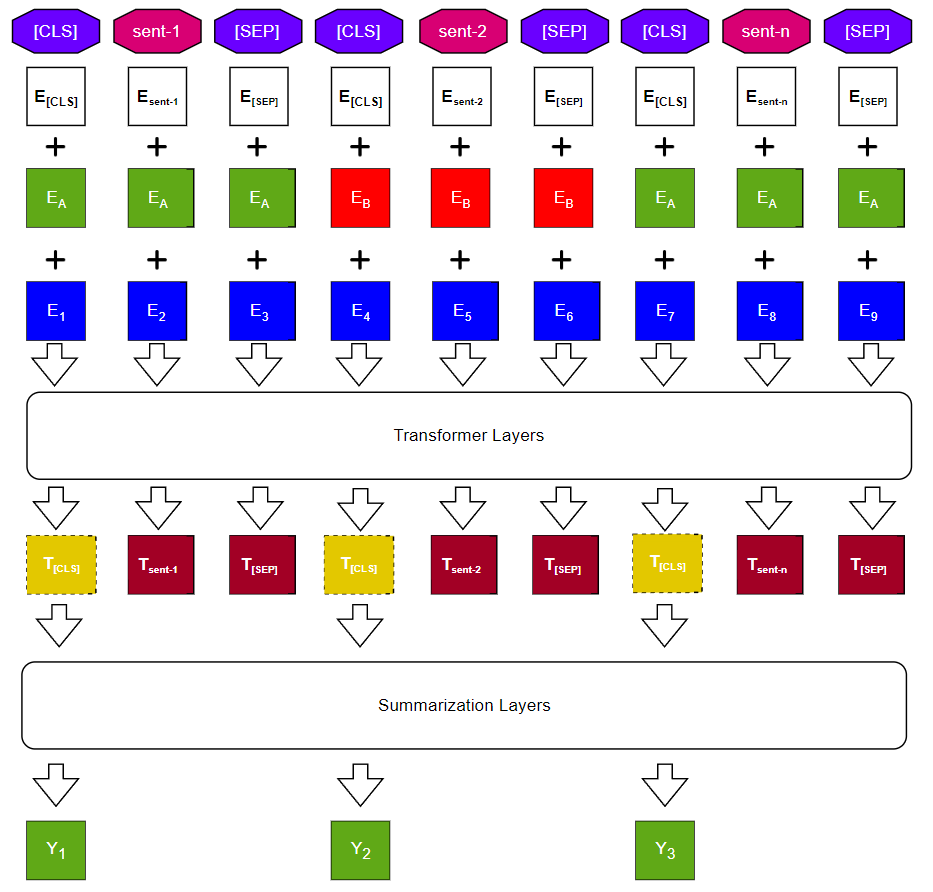}
    \caption{Overview for the architecture of the SAraBERT model}
    \label{fig:sarabert}
\end{figure}

Let $d$ denote a document containing several sentences $[sent_1,\ sent_2,\\ \ldots,\ sent_m]$, where $sent_i$ is the $i^{th}$ sentence in the document. Extractive summarization can be defined as the task of assigning a label $y_i\in\{0,1\}$ to each $sent_i$, indicating whether the sentence should be included in the summary or not. Thus, if $y_i=1$ then the sentence is considered in the summary due to its importance in terms of document content. BERT as shown in Figure[\ref{fig:bert}] is considered the model of choice because it showed better performance compared to other NLP statement embedding algorithms. To keep the original BERT pre-training objective, AraBERT uses Masked Language Modeling (MLM) to improve the pre-training process by letting the model predict the entire word rather than receiving clues from part of the word. In addition, it also uses Next Sentence Prediction (NSP) to help the model understand the relationships between sentences \cite{antoun2020arabert}. To achieve sentence ranking in the document based on content importance, we will apply modifications to the embedding layers as shown in Figure[\ref{fig:sarabert}] to highlight sentence endpoints for the model to rank each sentence independently.

\textbf{Encoding Multiple Sentences}: We insert $[CLS]$ token before each sentence and a $[SEP]$ token after each sentence. The $[CLS]$ is used as a symbol to represent the start of a sequence and the $[SEP]$ indicated the end of that sequence. Inserting multiple $[CLS]$ tokens help in specifying the boundaries of each sentence.

\textbf{Interval Segment Embedding}: We add another layer of embedding to distinguish multiple sentences within a document. So, for $sent_i$ we assign a segment embedding $E_A$ or $E_B$ based on whether it’s even or odd. For example, For $ [sent_1,\ sent_2,\ sent_3, sent_4,\ sent_5] $ the interval segment layer assign $[E_A,\ E_B,\ E_A, E_B, E_A]$.
The vector $T_i$ that corresponds to the $i^{th}$ $[CLS]$ token, will be used as the representation for $sent_i$.
The sequence on top is the input document, followed by the summation of three kinds of embeddings for each token. The summed vectors are used as input embeddings to the transformer layers, generating contextual vectors for each token $T_i$.
The difference between SAraBERT and the model in Figure[\ref{fig:bert}] is the insertion of [CLS] tokens before each sentence, and alternating the sentence segmenting embedding values along with the addition of a summarization layer that derives a sentence score from the [CLS] token embeddings.
\begin{algorithm}[H]
\caption{Summarization Model}\label{alg:cap}
\hspace*{\algorithmicindent}\textbf{Input}: Text \\
\hspace*{\algorithmicindent}\textbf{Output} : Sentence Scores
\begin{algorithmic}
\State $tokens  \gets   tokenize(\text{Text})$
\State $sentences   \gets   sentence\_segmentation(tokens)$
\State $tokenID  \gets array\left[|sentences|\right]\left[max_{sent\in sentences}(|sent|)\right]$
\State $segmentID \gets array\left[|sentences|\right]\left[max_{sent\in sentences}(|sent|)\right] $
\For{$i=1\text{ to }\left|sentences \right|$}
        \State sentences[i] $\gets \left\{ \text{"[CLS]"},\text{sentences[i]},\text{"[SEP]"} \right\} $
        \For{$j=1\text{ to }\left|sentences[i] \right|$}
            \State tokenID[i][j]$\gets$ getTokenID(sentences[i][j])  
            \If{$i$ is Even}
            \State segmentID[i][j]$\gets 0$
            \Else
            \State segmentID[i][j]$\gets 1$
            \EndIf
        \EndFor
\EndFor
\State bertEmbedding $\gets$ AraBERT(tokenID,segmentID) 
\State sentence\_embeddings $\gets$ get\_CLS\_embeddings(bertEmbedding)
\State sentence\_scores$\gets$ encode(sentence\_embeddings)
\end{algorithmic}
\end{algorithm}
After obtaining the embeddings of CLS tokens of each sentence, the tokens will get fed into an encoder $\hat{Y}=H(T)$ where $H$ is the encoding model, $T$ is a vector of CLS embeddings, and $\hat{Y}$is a scoring vector with $\hat{y_i}$ representing the score of the $i^{th}$ sentence. The model $H$ has been interchanged between Multi-Layer Perceptrons (MLP), Recurent Neural Network (RNN), and Transformers to compare the different scoring criteria each encoder have. Results are shown in section \ref{result}

\subsection{Siamese Semantic Similarity (SSS)}\label{SSS}
\par ROUGE \cite{lin2004rouge} has been used as a metric for determining the quality of a summary by comparing a candidate summary (generated by machine) to a reference summary (generated by humans). The way it does this projects only the level of similarity on the syntactic level without the coverage of the context. Summaries may cover a large portion of the original context but with fewer words (and more verbose) which Rouge can't keep track of. Language models have the ability to map the context of a passage into a fixed size vector. We can compute the cosine similarity of the 2 embeddings to obtain a similarity measure at the semantic level.

\begin{table}[H]
\centering
\caption{An example comparing the ROUGE score and Cosine Similarity}
\begin{tabular}{|l l|l l|} 
\hline
Reference  & \RL{علي يأكل الطعام في الليل} & ROUGE & Cosine Similarity                           \\ 
\hline
Candidate~ & \RL{هو يتناول الطعام مساء}  & 0.0   & 0.9074  \\
\hline
\end{tabular}

\end{table}

Since cosine similarity treats all dimensions equally in the semantic space, and 2 vector points might overlap yet stay very distant from each other, the distance should be added into the evaluation, since the smaller the value is the higher the score should be, we insert the distance difference in the denominator. Moreover, we will wrap it with a square root to slow the growth speed of the value as the difference increase (this will later become helpful in interpreting the computed values).

\begin{equation}
\frac{cos\_sim(cand_{embedding},ref_{embedding})}{\sqrt{||cand_{embedding}-ref_{embedding}||_2+1}}    
\end{equation}

A value of 1 was added because the distance difference can be 0, thus to eliminate the possibility of a division by zero error.

\begin{table}[h]
\centering
\caption{An example showing the cosine similarity and L2 norm when comparing a reference to predictions}
\begin{tabular}{|l l|l l|} 
\hline
 Reference  & \small{\RL{تناول الولد تفاحة}} & Cos\_Sim & L2 norm\\ 
\hline
Candidate$_1$ &	\small{\RL{أكل الطفل قطة}}	&	0.873	&	5.989	\\
\hline
Candidate$_2$ &	\small{\RL{أكل الطفل فاكهة}}	&	0.867	&	5.686	\\
\hline
\end{tabular}
\end{table}

Moreover, ROUGE is still an important component of the formula to keep an eye on the grammar. An average of ROUGE 1, ROUGE 2, and ROUGE L is used.
\begin{equation}
    \frac{cos\_sim(cand_{embedding},ref_{embedding})\cdot rouge(cand,ref)}{\sqrt{||cand_{embedding}-ref_{embedding}||_2+1}}
\end{equation}

Finally, frequency should be considered. Two sentences can be similar in context yet one being redundant more than the other.

\begin{table}[h]
\centering
\caption{An example showing the cosine similarity, ROUGE score, L2 norm, and frequency difference when comparing a reference to predictions}
\begin{tabular}{|l|l l l l|} 
\hline
\small{\RL{تناول الولد تفاحة}} & Cos\_Sim & Rouge & L2 norm & FreqDiff\\ 

\hline
\small{\RL{أكل الطفل تفاحة}}		&	0.957	&	0.222	&	3.637	&	1.000	\\
\hline
\begin{tabular}[c]{@{}l@{}} \small{\RL{تناول تناول \\الولد تفاحة تفاحة }}\end{tabular} &	0.930	&	0.889	&	5.005	&	1.667	\\
\hline
\end{tabular}
\end{table}

Thus a penalty should be added on the difference in frequency distribution among words where we divide the the number of unique words over the number of total words. This step was added because an embedding does not take redundancy and coverage into consideration which can affect level of similarity for tasks such as summarization. The final Formula can be seen in Algorithm[\ref{Siam}]
\begin{table}[H]
    \centering
    \caption{ interpretation of SSS scores}
    \label{tab:my_label}
    \begin{tabular}{|c|l| }
    \hline
    SSS score & Interpretation\\
    \hline
         0-0.1 &  no clear resemblance or no similarity at all\\ 
         0.1-0.2 & the gist is clear, but not all context is covered\\
         0.2-0.5 & context is significantly covered using different words\\
         $\geq$0.5 & exact replica with slight modifications\\
    \hline
    \end{tabular}
\end{table}

\begin{algorithm}[H]
\caption{Semantic Textual Similarity}\label{Siam}
\hspace*{\algorithmicindent}\textbf{Input}: Candidate, Reference \\
\hspace*{\algorithmicindent}\textbf{Output} : SimilarityScore
\begin{algorithmic}
\State $rouge_1 \gets rouge_1(\text{Candidate},\text{Reference})$
\State $rouge_2 \gets rouge_2(\text{Candidate},\text{Reference})$
\State $rouge_L \gets rouge_L(\text{Candidate},\text{Reference})$
\State rouge $\gets \frac{rouge_1+rouge_2+rouge_L}{3}$
\State $embedding_1 \gets AraBERT.encode(\text{Candidate})$
\State $embedding_2 \gets AraBERT.encode(\text{Reference})$
\State $cosine\_similarity \gets \frac{embedding_1\cdot embedding_1} {\left\| embedding_1 \right\|_2\times \left\| embedding_2 \right\|_2}$
\State $DistDiff \gets \left\| embedding_1 - embedding_2 \right\|_2$
\State $FreqDiff_c \gets \frac{unique\_words(Candidate)} {total\_words(Candidate)} $
\State $FreqDiff_r \gets \frac{ unique\_words(Reference)} { total\_words(Reference) } $
\State $FreqDiff \gets \frac{min(FreqDiff_r,FreqDiff_c)}{max(FreqDiff_r,FreqDiff_c)}$
\State SimilarityScore $ \gets \frac{ \text{cosine\_similarity}\times \text{rouge} \times FreqDiff }{
\sqrt{DistDiff+1}}$

\end{algorithmic}
\end{algorithm}

\section{Experiment}\label{experiment}

\subsection{Dataset}
\subsubsection{CNN/ Daily Mail}
We used the CNN / DailyMail dataset to train SAraBERT, it is an English dataset containing over 300,000 unique news articles written by CNN and DailyMail journalists. The dataset can be used to train a model for the task of extractive and abstractive summarization. Each document instance from the dataset consists of 3 components:
\begin{enumerate}
    \item \textbf{id}: A string containing the hexadecimal SHA1 hash of the URL from which the story was obtained. 
    \item \textbf{Article}: A string containing the body of a news article.
    \item \textbf{Highlights}: A string containing the highlights of the article written by the author of the article. 
\end{enumerate}

Since our model should be trained on an Arabic dataset, we examined state-of-the-art machine translation model mbart \cite{tang2020multilingual} along with Google-Trans\footnote{\href{https://pypi.org/project/googletrans/}{pypi.org/googletrans}} service. We evaluated each of the models on the Opus dataset \cite{tang2020multilingual}. Results  are shown in Table[\ref{translators}]. We can see competitive results from both translators, so the tie breaker was the duration required to translate a document.
\begin{table}[H]
\centering
\caption{Translation Experiment Results}

\begin{tabular}{|l|l|l|l|} 
\hline
& \textbf{BLEU} & \textbf{METEOR} & \begin{tabular}[c]{@{}l@{}}\textbf{Duration}\\\textbf{(Docs/minute)}\end{tabular}  \\ 
\hline
\textbf{GoogleTrans} & 0.413 $\pm$ 0.386 & 0.24            & 30                                                                                             \\ 
\hline
\textbf{mBart}       & 0.413 $\pm$ 0.382   & 0.25            & 20                                                                                             \\
\hline
\end{tabular}
\label{translators}
\end{table}

\begin{figure}[H]
\centering
\includegraphics[width=\columnwidth]{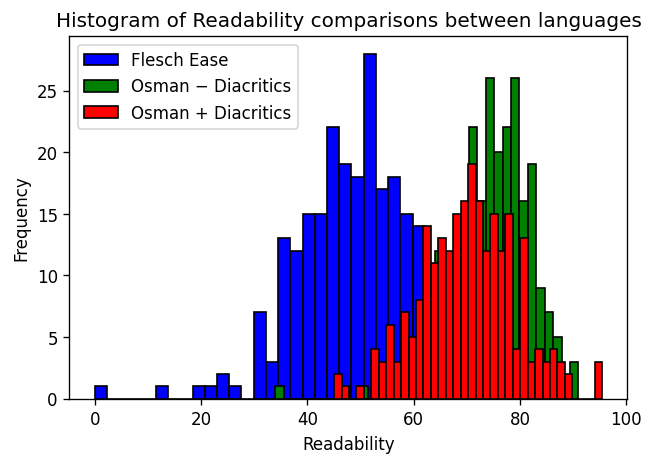}
\caption{Histogram of readability comparisons between english and arabic with and without diacritics}
\label{osmanHist}
\end{figure}

We also generated diacriticisation for the Arabic documents \cite{Shakkala} to see whether the diacritics have an effect on the readability level or not.
Table[\ref{osman}] and Figure[\ref{osmanHist}] shows that the documents containing diacritics have closer readability estimations to English than documents without diacritics but it will not be a problem since \cite{el2016osman} explained that this behavior is normal and what is important is that the correlation between Flesch and OSMAN (without diacritics) is high.
\begin{figure}[h!]
\centering
\includegraphics[width=\columnwidth]{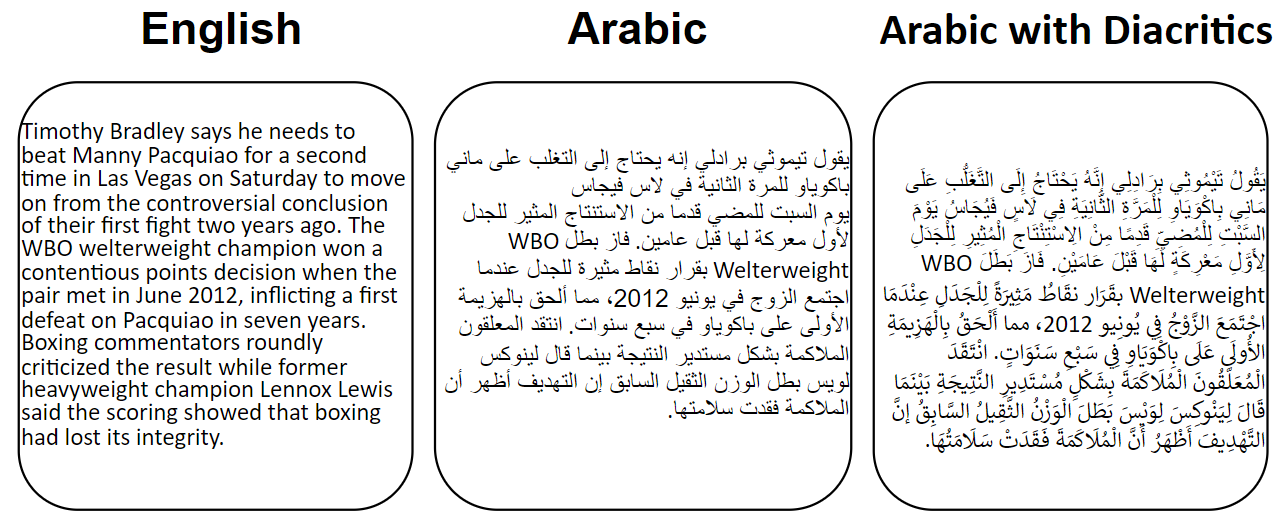}
\caption{Sample English passage translated to Arabic and Diactritized}
\end{figure}\\
We investigated if translation will affect the readability of a document and whether large difference between the English and Arabic readabilities of the original and translated documents will influence our model. After experimenting on 1000 translated samples, and computing the ROUGE score of the resulting summaries along with the readability of the original and translated samples, results have shown that the level of readability does not affect the model's output and there is no correlation between readability difference and ROUGE. More details are provided in Appendix[\ref{translationQuality}].

\begin{table}[h!]
\centering
\caption{Average Readability Measures for Arabic}
\captionsetup{}
\begin{tabular}{|l|l|} 
\hline
 Metric & Value (Mean $\pm$ std)  \\ 
\hline
OSMAN - Diacritics &   78.45$\pm$5.18\\
\hline
OSMAN + Diacritics &   70.28$\pm$8.95\\
\hline
\end{tabular}

\label{flesch}
\label{osman}
\end{table}

\section{Results}\label{result}
Results are reported in Table[\ref{resultTable}]. The 3 metrics used for evaluation are BLEU (Average of the 4 BLEU evaluations over uni-,bi-,tri-, and quad-grams), ROUGE (Average between F1-Scores of ROUGE-1,ROUGE-2, and ROUGE-L), and the Siamese Similarity metric. Oracle Score represents the evaluation of human summaries that will be used for comparison with machine summaries. "SAraBert+RNN" was the best performing model based on the 3 evaluation metrics. Sample summaries given by the different models can be found in the Appendix[\ref{Sample Summaries}].

\begin{table*}[htpb!]
\caption{BLEU scores over Kalimat Dataset}
\centering
\begin{tabular}{|l|llll|lll|} 
\hline
\textbf{Model}               & \textbf{bleu\_1} & \textbf{bleu\_2} & \textbf{bleu\_3} & \textbf{bleu\_4} & \textbf{avg} & \textbf{min}& \textbf{max} \\ 
\hline
SAraBert + Classifier        & 0.408             & 0.385            & 0.38            & 0.376            & 0.387 &0.0 & 0.99        \\ 
\hline
SAraBert + RNN               & \textbf{0.474}            & \textbf{0.45}            & \textbf{0.445}            & \textbf{0.441}            & \textbf{0.452}     &0.0&0.99     \\ 
\hline
\begin{tabular}[c]{@{}l@{}}SAraBert  \\
+ Transformer\end{tabular}       & 0.391            & 0.368           & 0.364            & 0.36            & 0.371 &0.0&0.99         \\ 
\hline
\begin{tabular}[c]{@{}l@{}}AraBert Embeddings \\
+ K-Means\end{tabular} & 0.284            & 0.254            & 0.249            & 0.246            & 0.258 &0.0 &0.991          \\ 
\hline
AraVec + K-Means             & 0.345            & 0.334            & 0.331            & 0.328            & 0.335 &0.0&0.973         \\ 
\hline
Bag of Words                 & 0.361            & 0.351            & 0.348            & 0.345            & 0.352    &0.0&0.98      \\
\hline
\end{tabular}
\end{table*}

\begin{table*}[htpb!]
\centering
\caption{ROUGE scores over Kalimat Dataset}
\begin{tabular}{|l|lll|lll|} 
\hline
\textbf{Model}               & \textbf{rouge\_1} & \textbf{rouge\_2} & \textbf{rouge\_L} & \textbf{avg} & \textbf{min} & \textbf{max}  \\ 
\hline
SAraBert + Classifier        & 0.529             & 0.466             & 0.51              & 0.502        &     0.008    & 0.99         \\ 
\hline
SAraBert + RNN               & \textbf{0.588}             & \textbf{0.524}             & \textbf{0.568}             & \textbf{0.56}        &   0.008    &   0.99   \\ 
\hline
\begin{tabular}[c]{@{}l@{}}SAraBert  \\
+ Transformer\end{tabular}       & 0.512             & 0.444             & 0.491             & 0.482   &   0.017       & 0.99    \\ 
\hline
\begin{tabular}[c]{@{}l@{}}AraBert Embeddings \\
+ K-Means\end{tabular} & 0.445             & 0.366             & 0.427             & 0.414        &   0.008    &   1.0            \\ 
\hline
AraVec + K-Means             & 0.545             & 0.49              & 0.535             & 0.523        &        0.0      &      1.1         \\ 
\hline
Bag of Words                 & 0.47             & 0.415             & 0.46             & 0.448        &         0.0      &      1.1        \\
\hline
\end{tabular}
\end{table*}

\begin{table*}
\centering
\caption{Results of different models under several evaluation metrics}
\begin{tabular}{|l|l|l|l|l|l|}
\hline
\textbf{Model}                                      & \textbf{BLEU} & \textbf{ROUGE} & \textbf{\begin{tabular}[c]{@{}l@{}}Siamese \\ Similarity\end{tabular}}& \textbf{BS}& \textbf{TS-SS} \\ \hline
Oracle Score                                        & 0.532 ± 0.27         & 0.699 ± 0.23              & 0.326 ± 0.3 &0.83   &0.003
             \\ \hline
SAraBert + Classifier                               & 0.387 ± 0.3         & 0.502 ± 0.31             & 0.211 ± 0.158  & 0.76 &0.018\\ \hline
SAraBert + RNN                                      & \textbf{0.452} ± 0.2& \textbf{0.560} ± 0.26    & \textbf{0.282} ± 0.18&\textbf{0.79} &0.017
 \\ \hline
\begin{tabular}[c]{@{}l@{}}SAraBert  \\
+ Transformer\end{tabular}                          & 0.371 ± 0.3         & 0.482 ± 0.31             & 0.240 ± 0.17   &0.77 &0.019   \\ \hline
\begin{tabular}[c]{@{}l@{}}AraBert Embeddings \\
+ K-Means\end{tabular}                              & 0.258 ± 0.21         & 0.414 ± 0.25             & 0.216 ± 0.1  &0.76 &0.032    \\ \hline
AraVec + K-Means                                    & 0.335 ± 0.28         & 0.523 ± 0.28             & 0.191 ± 0.18  &0.74  &0.039     \\ \hline
Bag of Words                                        & 0.352 ± 0.35          & 0.448 ± 0.36             & 0.217 ± 0.28  &0.76  &0.111   \\ \hline
\end{tabular}
\label{resultTable}
\end{table*}

\begin{figure}
    \centering
    \includegraphics[width=\columnwidth]{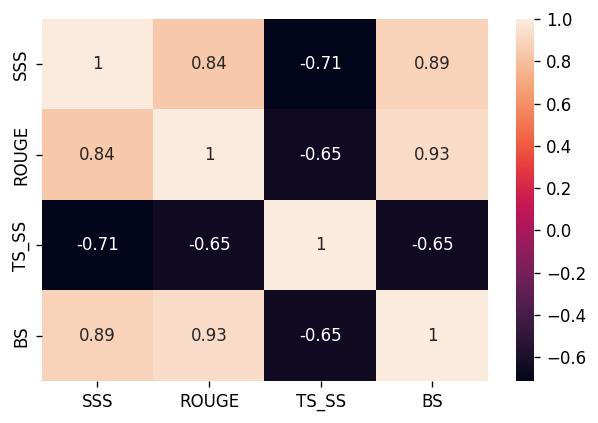}
    \caption{Pearson Correlation between the similarity metrics}
    \label{fig:correlation}
    \captionsetup{justification=centering}
\end{figure}

\section{Discussion}
It appears that the usage of BiLSTM encoder did better than the transformer encoder, this could be due to the need of transformers for huge amount of data to train all its attention heads and layers. Another interesting thing to notice is that Bag of Words method got a high score on the Siamese Similarity distinguishably higher than the other classical approaches. One problem remains is the incapability of feeding very large documents into the models to obtain a single global lookup on the document instead of segmenting the document and loosing linked contexts between the trimmed passages. Future work will focus on the SSS metric, where the cosine similarity should give more importance/weight/attention to specific context features.

\section{Conclusion}\label{conclusion}
Text summarization is one of the important areas of research in NLP, as textual data keep increasing each day. The proposed model "\emph{SAraBert}" works on the summarization task for MSA NLP. We applied experiments on a large-scale translated dataset and found that SAraBert with RNN encoding layers can achieve the best performance. We also created a similarity metric that evaluated the similarity of 2 documents on the semantic level and not just the syntactic level.

\clearpage
\section*{Appendix}
\subsection{Siamese Semantic Similarity Experiments}
The metric that was discussed is Section[\ref{SSS}] had to be tested for different cases to visualise and evaluate the overall performance of this metric.

We applied SSS on a set of candidates shown in Table[\ref{SSScandidates}] to the reference found at the first row of that table. The results of SSS and BertScore can be found in Table[\ref{SSSresults}] along with the values required to compute SSS. The correlation between BertScore and SSS is $0.97$ showing that both metrics behave very similar but SSS tends to be more strict with its evaluations.

{\tiny
\begin{table*}[t]
\caption{List of a reference sentence (1) and set of candidate sentences(2 to 19) }

\centering
\begin{tabular}{|l|l|}
\hline
\multicolumn{1}{|l|}{ID} & \multicolumn{1}{l|}{Text}       \\ 
0                        & \RL {أكل التلميذ سامي طعامه في المدرسة صباحا يوم الثلاثاء. طعامه كان تفاحة}\\
1                        & \RL {هرب}\\
2                        &\RL {أكل }\\
3           			 &\RL{أكل التلميذ سامي طعامه في المدرسة صباحا يوم الثلاثاء. طعامه كان تفاحة}\\
4           			 &\RL{تناول التلميذ سامي طعامه في المدرسة صباحا يوم الثلاثاء. طعامه كان تفاحة}\\
5           			&\RL{تناول سامي طعامه في المدرسة صباحا يوم الثلاثاء. طعامه كان تفاحة}\\
6                        &\RL {أكل التلميذ سامي طعامه في المدرسة صباحا يوم الثلاثاء. طعامه كان تفاحة }\\
7                        &\RL {أكل سامي }\\
8                        &\RL {أكل سامي تفاحة}\\
9                        &\RL{أكل سامي دجاجة}\\
10                       &\RL{أكل سامي طعاما}\\
11                       &\RL{أكل التفاح سامي}\\
12                       &\RL {أكلت سامية طعامها}\\
13                       &\RL{في المدرسة تناول سامي وجبة تفاح}\\
14                       &\RL {في المدرسة أكل سامي}\\
15                       &\RL {نهار الثلاثاء صباحا أكل سامي تفاحة}\\
16                       &\RL {فازت الممثلة جيسيكا بجائزة اوسكار}\\
17                       &\RL{كان يصوم معها حتى فهم أن ذلك يعني أنه لا يستطيع الأكل}\\
18                       &\RL{اسمح لي أن أساعدك في أمتعتك }\\
19                       &\RL{استيقظ سامي صباح يوم الثلاثاء متحمسا للذهاب الى المدرسة لانه يريد ان يتناول تفاحة هناك }\\
20                       &\RL{أكل سامي تفاحة أكل سامي تفاحة أكل سامي تفاحة أكل سامي تفاحة أكل سامي تفاحة}\\
21                       &
\begin{tabular}[c]{@{}l@{}}
\RL{ أكل أكل  أكل أكل أكل  أكل التلميذ سامي سامي سامي طعامه المدرسة المدرسة في المدرسة المدرسة}
 \\ 
\RL{  صباحا يوم الثلاثاء يوم يوم يوم. طعامه كان تفاحة تفاحة تفاحة تفاحة تفاحة}
\end{tabular}
\\\hline
\end{tabular}
\label{SSScandidates}
\end{table*}
}
\begin{table}[H]
\caption{ Results of computing SSS and BertScore (BS) along with other required computations for the candidates in Table[\ref{SSScandidates}] linked by ID }
\begin{tabular}{c|c|c|c|c|c|c}%
\bfseries ID & \bfseries Cosine& \bfseries ROUGE&\bfseries norm$_2$&\bfseries FD&\bfseries SSS &\bfseries BS
\csvreader[head to column names]{SSSeval.csv}{}
{\\\hline\index & \CosineSimilarity&\rouge&\L&\fd&\sss&\bertscore}
\end{tabular}

\label{SSSresults}
\end{table}
\subsection{Sample Summaries form SaraBert}\label{Sample Summaries}
This section will present sample summaries extracted via SaraBert, highlights are used to visualize the selected top 3 sentences selected to be considered as the most informative sentences.
\begin{figure}[H]
    \centering
    \includegraphics[width=\columnwidth]{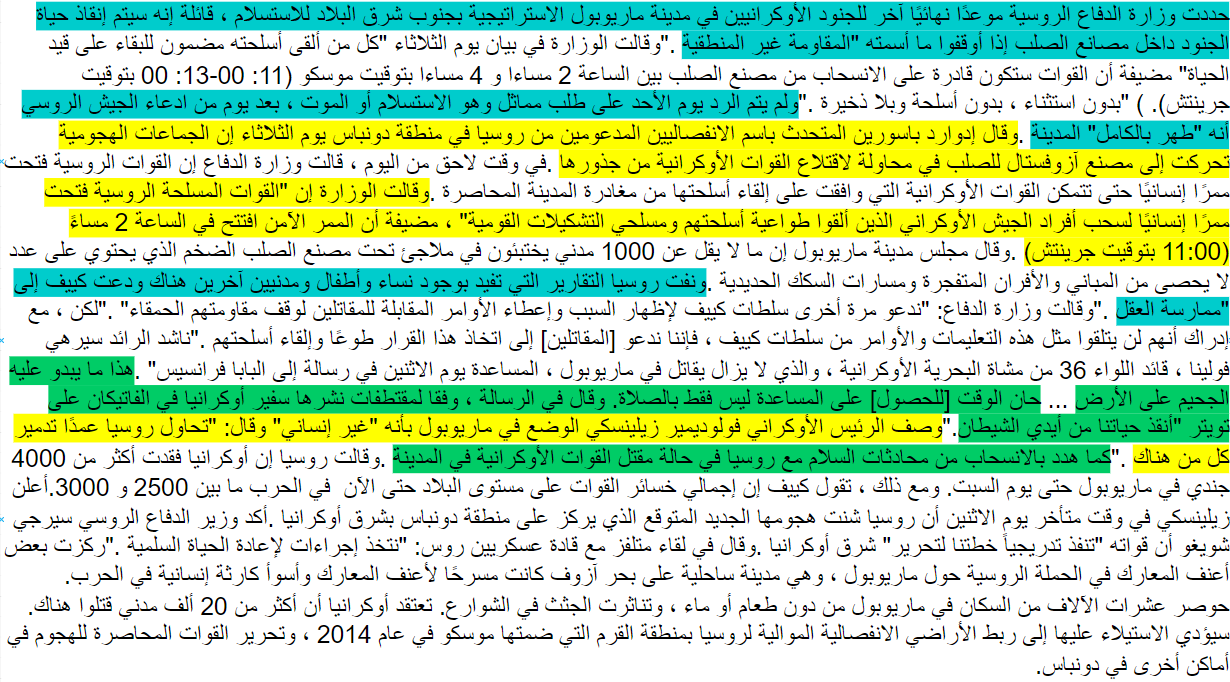}
    \caption{Sample SaraBert summarization. Yellow highlighting represents summary extraction using MLP as encoder, Green represents RNN encoder, and Blue represents Transformer}
    \label{fig:sample1}
\end{figure}


\begin{figure}[H]
    \centering
    \includegraphics[width=\columnwidth]{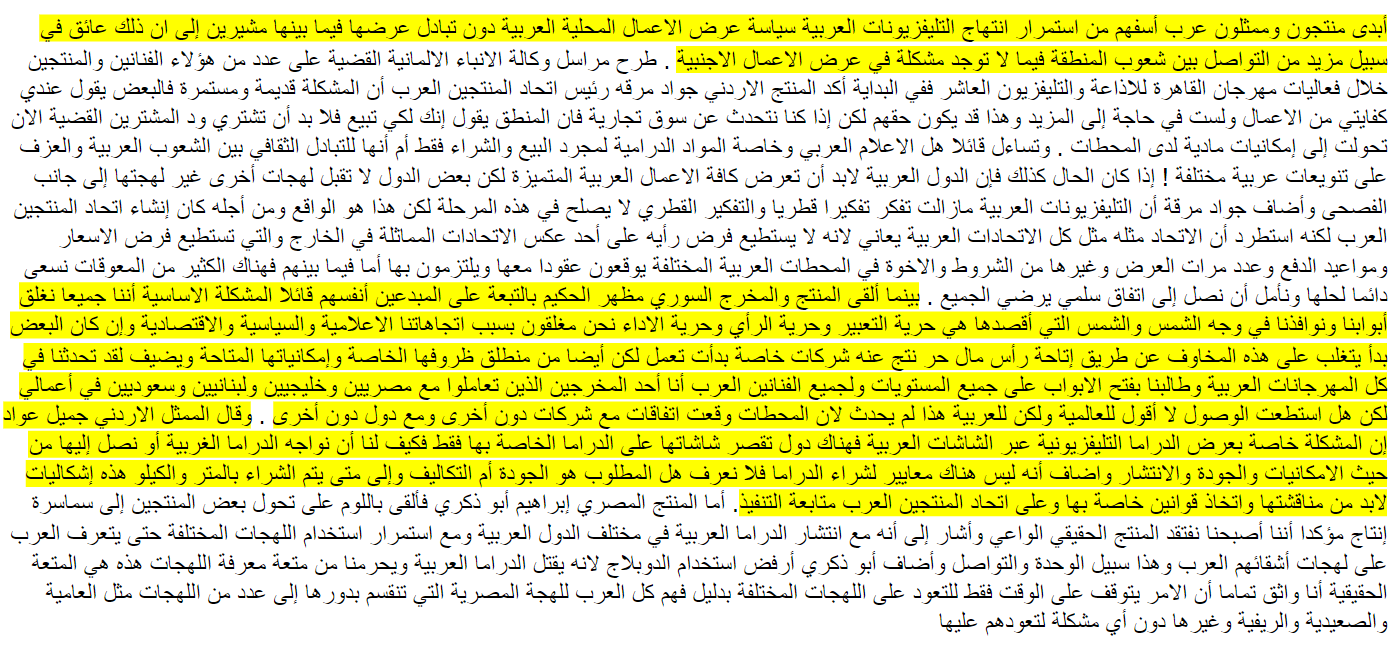}
    \caption{Sample SaraBert+BiLSTM of a good summarization}
    \label{fig:sample good}
\end{figure}

\begin{figure}[H]
    \centering
    \includegraphics[width=\columnwidth]{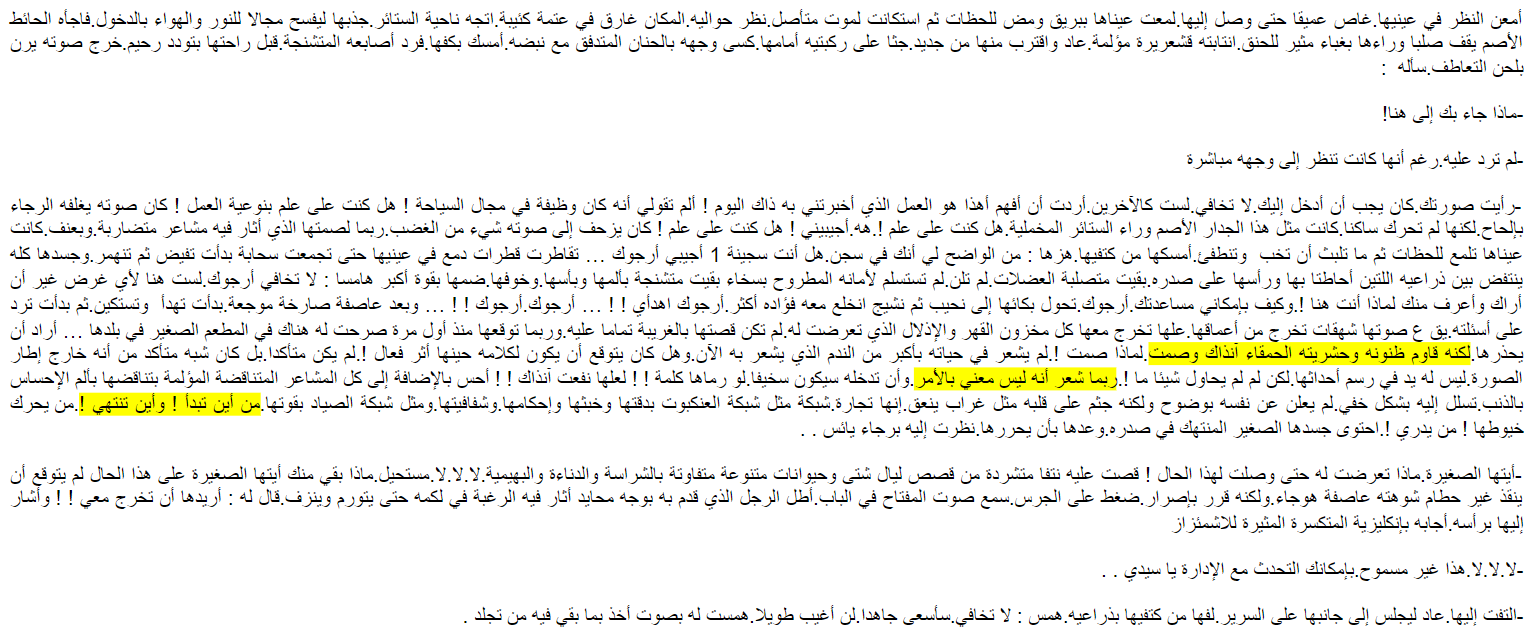}
    \caption{Sample SaraBert+BiLSTM of a bad summarization}
    \label{fig:sample bad}
\end{figure}

\subsection{Translation Quality}\label{translationQuality}
In this section we will present sample translations and discuss the quality and performance effect on the results.
\begin{figure}[H]
    \centering
    \includegraphics[width=\columnwidth]{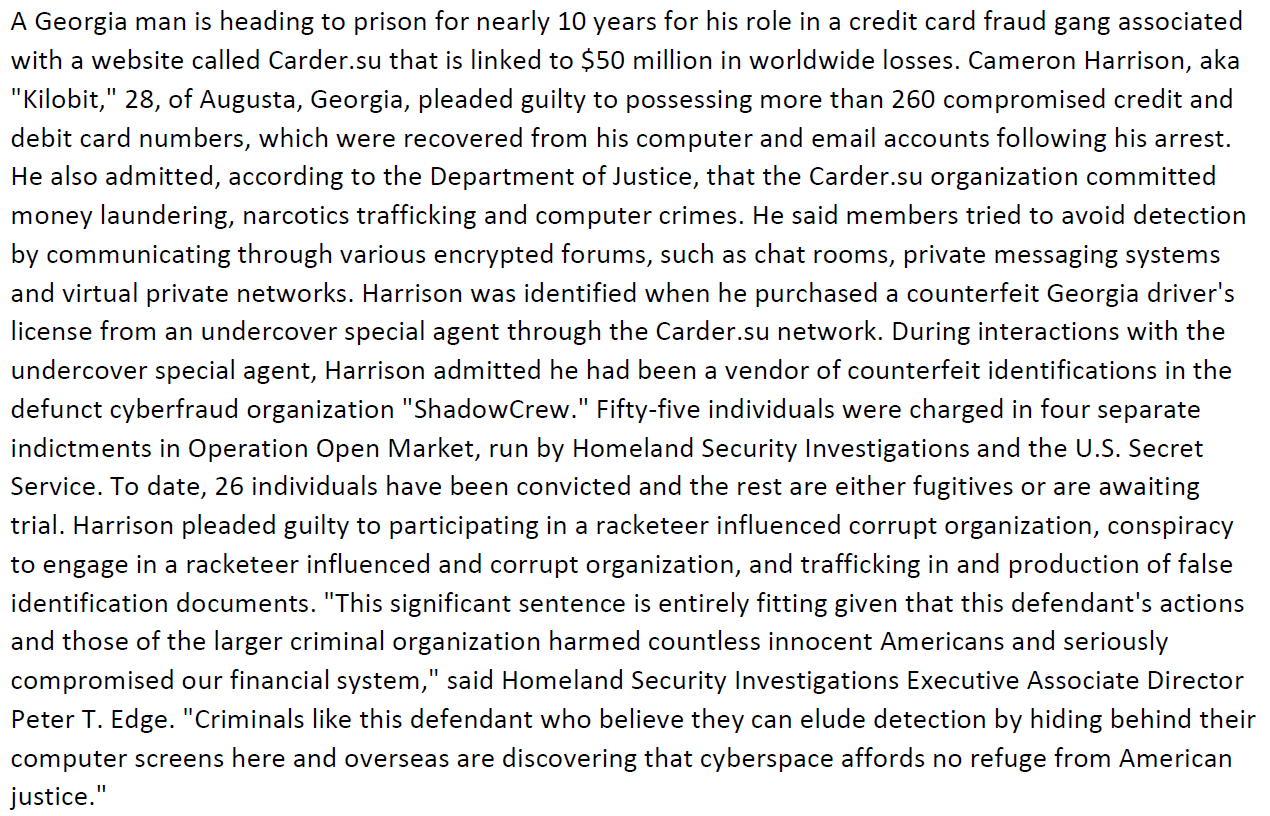}
    \caption{Sample English passage with Fleach score of 11.38 (Hard to read)}
    \label{fig:trans1}
\end{figure}
\begin{figure}[H]
    \centering
    \includegraphics[width=\columnwidth]{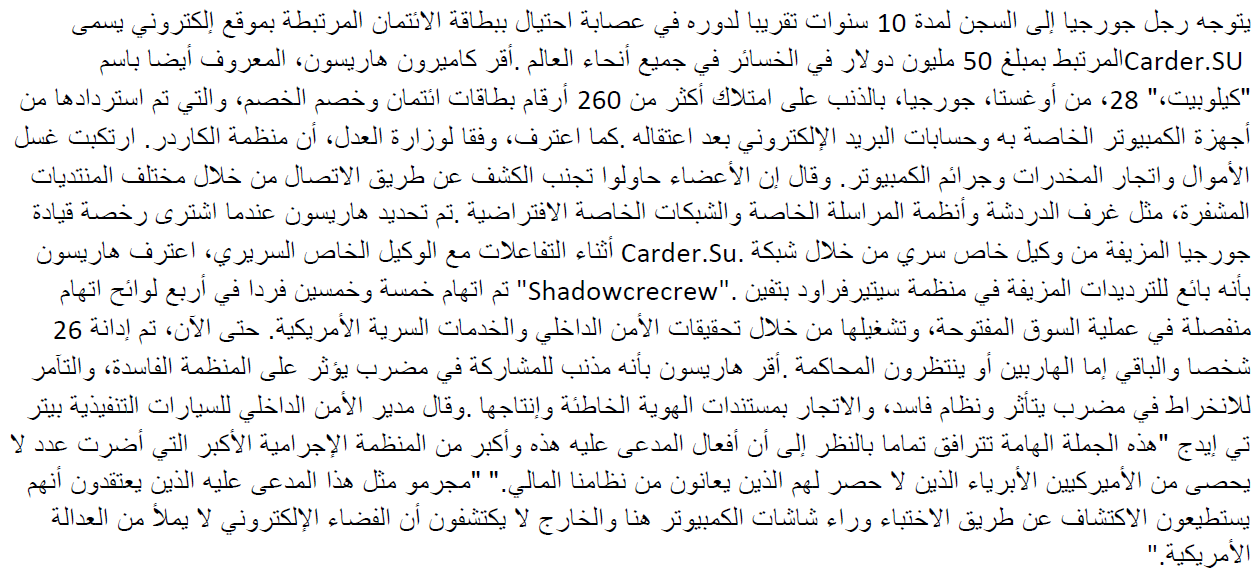}
    \caption{Sample Arabic passage (translation of Figure[\ref{fig:trans1}]) with Osman score of 66.73 (Slight hard to read)}
    \label{fig:trans2}
\end{figure}

\begin{figure}[H]
    \centering
    \includegraphics[width=\columnwidth]{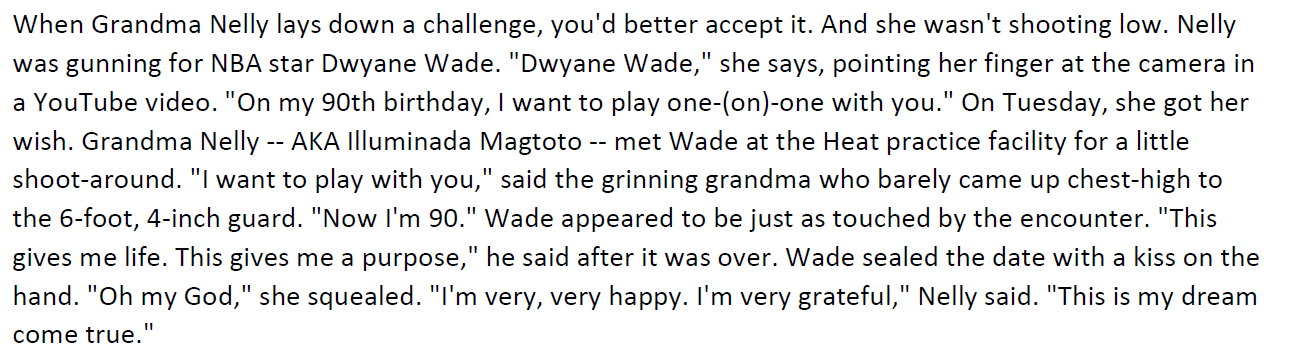}
    \caption{Sample English passage with Fleach score of 83.69 (Easy to read)}
    \label{fig:trans3}
\end{figure}
\begin{figure}[H]
    \centering
    \includegraphics[width=\columnwidth]{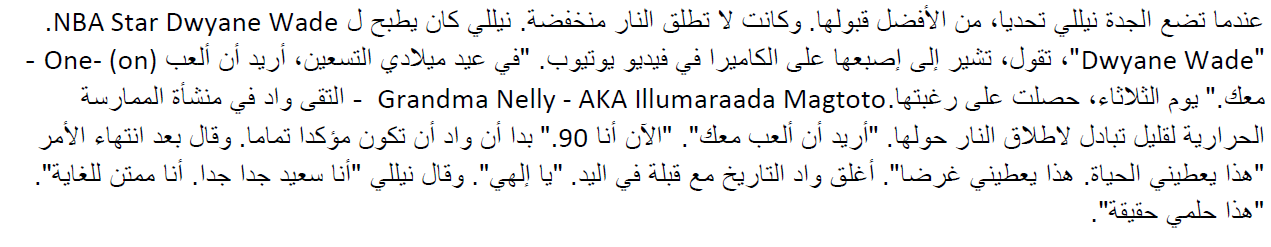}
    \caption{Sample Arabic passage (translation of Figure[\ref{fig:trans3}]) with Osman score of 88.01 (Easy to read)}
    \label{fig:trans4}
\end{figure}

The translations have shown no effect on the trained model, where the readability variation had no link with ROUGE results. We sampled 1000 documents and computed their respective readability metrics in Arabic and English (Osman and Flesch) along with the ROUGE value of the generated summary. 
By looking at table[\ref{CorrelationDoc}], it can be observed that no readability metric highly affects the ROUGE evaluation.

\begin{table}[H]
\caption{Average value for documents computed over 1000 sample documents}
\centering
\begin{tabular}{|l|l|}
\hline
Flesch              & 59.93 $\pm$ 10.13  \\
Osman               & 74.88 $\pm$ 7.79   \\
Rouge-AR            & 0.41 $\pm$ 0.14   \\
Readability-Diff    & 14.95 $\pm$ 6.55   \\
\hline
\end{tabular}
\end{table}
\begin{table}
\caption{Correlation between the different features}

\centering
\begin{tabular}{|l|cccc|}
\hline
Index             & Flesch & Osman & Rouge-AR & Readability-Diff  \\
\hline
Flesch            & 1.0    & 0.76  & -0.12     & -0.64              \\
Osman             & 0.76   & 1.0   & -0.04     & 0.01               \\
rouge-AR         & -0.12  & -0.04 & 1.0       & 0.14               \\
Readability-Diff & -0.64  & 0.01  & 0.14      & 1.0               \\
\hline
\end{tabular}
\label{CorrelationDoc}
\end{table}

\clearpage
\bibliographystyle{ieeetr}
\bibliography{references}

@inproceedings{Hajj2013,
    author = {Hajj,Nadine and Awad,Mariette} ,
    title = {Weighted entropy cortical algorithms for isolated Arabic speech recognition},
    year = {2013}, 
    publisher={IEEE},

}

@article{jing2002using,
  title={Using hidden markov modeling to decompose human-written summaries},
  author={Jing, Hongyan},
  journal={Computational linguistics},
  volume={28},
  number={4},
  pages={527--543},
  year={2002},
  publisher={MIT Press One Rogers Street, Cambridge, MA 02142-1209, USA journals-info~…}
}

@article{al2016automatic,
  title={Automatic Arabic text summarization: a survey},
  author={Al-Saleh, Asma Bader and Menai, Mohamed El Bachir},
  journal={Artificial Intelligence Review},
  volume={45},
  number={2},
  pages={203--234},
  year={2016},
  publisher={Springer}
}

@article{al2015multi,
  title={Multi-document arabic text summarization},
  author={Al Harazin, Karim S},
  year={2015},
  publisher={الجامعة الإسلامية-غزة}
}

@article{patil2004statistical,
  title={A statistical approach for document summarization},
  author={Patil, Vishal and Krishnamoorthy, Mahalakshmy and Oke, Parag and Kiruthika, M},
  journal={Department of Computer Engineering Fr. C. Rodrigues Institute of Technology, Vashi, Navi Mumbai, Maharashtra, India},
  year={2004}
}

@article{alotaiby2012new,
  title={New approaches to automatic headline generation for Arabic documents},
  author={Alotaiby, Fahad and Foda, Salah and Alkharashi, Ibrahim},
  journal={Journal of Engineering and Computer Innovations},
  volume={3},
  number={1},
  pages={11--25},
  year={2012},
  publisher={Academic Journals}
}

@article{ferreira2013assessing,
  title={Assessing sentence scoring techniques for extractive text summarization},
  author={Ferreira, Rafael and de Souza Cabral, Luciano and Lins, Rafael Dueire and e Silva, Gabriel Pereira and Freitas, Fred and Cavalcanti, George DC and Lima, Rinaldo and Simske, Steven J and Favaro, Luciano},
  journal={Expert systems with applications},
  volume={40},
  number={14},
  pages={5755--5764},
  year={2013},
  publisher={Elsevier}
}

@inproceedings{al2009arabic,
  title={Arabic text summarization using aggregate similarity},
  author={Al-Radaideh, Q and Afif, Mohammad},
  booktitle={International Arab conference on information technology (ACIT2009), Yemen},
  year={2009}
}

@article{haboush2012arabic,
  title={Arabic text summarization model using clustering techniques},
  author={Haboush, Ahmad and Al-Zoubi, Maryam and Momani, Ahmad and Tarazi, Motassem},
  journal={World of Computer Science and Information Technology Journal (WCSIT) ISSN},
  pages={2221--0741},
  year={2012}
}

@article{el2014multi,
  title={Multi-topic multi-document summarizer},
  author={El-Ghannam, Fatma and El-Shishtawy, Tarek},
  journal={arXiv preprint arXiv:1401.0640},
  year={2014}
}

@inproceedings{fejer2014automatic,
  title={Automatic Arabic text summarization using clustering and keyphrase extraction},
  author={Fejer, Hamzah Noori and Omar, Nazlia},
  booktitle={Proceedings of the 6th International Conference on Information Technology and Multimedia},
  pages={293--298},
  year={2014},
  organization={IEEE}
}

@article{hewahi2012automatic,
  title={Automatic Arabic Text Summarization System (AATSS) Based on Semantic Features Extraction},
  author={Hewahi, Nabil M and Kwaik, Kathrein Abu},
  journal={International Journal of Technology Diffusion (IJTD)},
  volume={3},
  number={2},
  pages={12--27},
  year={2012},
  publisher={IGI Global}
}

@article{fattah2009ga,
  title={GA, MR, FFNN, PNN and GMM based models for automatic text summarization},
  author={Fattah, Mohamed Abdel and Ren, Fuji},
  journal={Computer Speech \& Language},
  volume={23},
  number={1},
  pages={126--144},
  year={2009},
  publisher={Elsevier}
}

@incollection{belkebir2015supervised,
  title={A supervised approach to arabic text summarization using adaboost},
  author={Belkebir, Riadh and Guessoum, Ahmed},
  booktitle={New contributions in information systems and technologies},
  pages={227--236},
  year={2015},
  publisher={Springer}
}

@article{al2018hybrid,
  title={A hybrid approach for arabic text summarization using domain knowledge and genetic algorithms},
  author={Al-Radaideh, Qasem A and Bataineh, Dareen Q},
  journal={Cognitive Computation},
  volume={10},
  number={4},
  pages={651--669},
  year={2018},
  publisher={Springer}
}

@incollection{nenkova2012survey,
  title={A survey of text summarization techniques},
  author={Nenkova, Ani and McKeown, Kathleen},
  booktitle={Mining text data},
  pages={43--76},
  year={2012},
  publisher={Springer}
}

@article{bahdanau2014neural,
  title={Neural machine translation by jointly learning to align and translate},
  author={Bahdanau, Dzmitry and Cho, Kyunghyun and Bengio, Yoshua},
  journal={arXiv preprint arXiv:1409.0473},
  year={2014}
}

@inproceedings{sutskever2014sequence,
  title={Sequence to sequence learning with neural networks},
  author={Sutskever, Ilya and Vinyals, Oriol and Le, Quoc V},
  booktitle={Advances in neural information processing systems},
  pages={3104--3112},
  year={2014}
}

@article{rush2015neural,
  title={A neural attention model for abstractive sentence summarization},
  author={Rush, Alexander M and Chopra, Sumit and Weston, Jason},
  journal={arXiv preprint arXiv:1509.00685},
  year={2015}
}

@article{nallapati2016classify,
  title={Classify or select: Neural architectures for extractive document summarization},
  author={Nallapati, Ramesh and Zhou, Bowen and Ma, Mingbo},
  journal={arXiv preprint arXiv:1611.04244},
  year={2016}
}

@inproceedings{chopra2016abstractive,
  title={Abstractive sentence summarization with attentive recurrent neural networks},
  author={Chopra, Sumit and Auli, Michael and Rush, Alexander M},
  booktitle={Proceedings of the 2016 Conference of the North American Chapter of the Association for Computational Linguistics: Human Language Technologies},
  pages={93--98},
  year={2016}
}

@article{al2020arabic,
  title={Arabic text summarization using deep learning approach},
  author={Al-Maleh, Molham and Desouki, Said},
  journal={Journal of Big Data},
  volume={7},
  number={1},
  pages={1--17},
  year={2020},
  publisher={Springer}
}

@article{antoun2020arabert,
  title={Arabert: Transformer-based model for arabic language understanding},
  author={Antoun, Wissam and Baly, Fady and Hajj, Hazem},
  journal={arXiv preprint arXiv:2003.00104},
  year={2020}
}

@article{abu2020arabic,
  title={Arabic Text Summarization Using AraBERT Model Using Extractive Text Summarization Approach},
  author={Abu Nada, Abdullah M and Alajrami, Eman and Al-Saqqa, Ahmed A and Abu-Naser, Samy S},
  year={2020}
}

@article{liu2019fine,
  title={Fine-tune BERT for extractive summarization},
  author={Liu, Yang},
  journal={arXiv preprint arXiv:1903.10318},
  year={2019}
}

@inproceedings{zalmout2017don,
  title={Don’t throw those morphological analyzers away just yet: Neural morphological disambiguation for Arabic},
  author={Zalmout, Nasser and Habash, Nizar},
  booktitle={Proceedings of the 2017 Conference on Empirical Methods in Natural Language Processing},
  pages={704--713},
  year={2017}
}

@inproceedings{lin2004rouge,
  title={Rouge: A package for automatic evaluation of summaries},
  author={Lin, Chin-Yew},
  booktitle={Text summarization branches out},
  pages={74--81},
  year={2004}
}

@inproceedings{el2011multi,
  title={Multi-document Arabic text summarisation},
  author={El-Haj, Mahmoud and Kruschwitz, Udo and Fox, Chris},
  booktitle={2011 3rd Computer Science and Electronic Engineering Conference (CEEC)},
  pages={40--44},
  year={2011},
  organization={IEEE}
}

@misc{inoue2021interplay,
      title={The Interplay of Variant, Size, and Task Type in Arabic Pre-trained Language Models}, 
      author={Go Inoue and Bashar Alhafni and Nurpeiis Baimukan and Houda Bouamor and Nizar Habash},
      year={2021},
      eprint={2103.06678},
      archivePrefix={arXiv},
      primaryClass={cs.CL}
}

@article{miller2019leveraging,
  title={Leveraging BERT for extractive text summarization on lectures},
  author={Miller, Derek},
  journal={arXiv preprint arXiv:1906.04165},
  year={2019}
}

@article{el2016osman,
  title={OSMAN: A novel Arabic readability metric},
  author={El-Haj, Mahmoud and Rayson, Paul Edward},
  year={2016},
  publisher={European Language Resources Association (ELRA)}
}

@misc{Shakkala,
  title={Shakkala, Arabic text vocalization},
  author={Barqawi, Zerrouki},
  url={https://github.com/Barqawiz/Shakkala},
  year={2017}
}

@article{tang2020multilingual,
  title={Multilingual translation with extensible multilingual pretraining and finetuning},
  author={Tang, Yuqing and Tran, Chau and Li, Xian and Chen, Peng-Jen and Goyal, Naman and Chaudhary, Vishrav and Gu, Jiatao and Fan, Angela},
  journal={arXiv preprint arXiv:2008.00401},
  year={2020}
}

@INPROCEEDINGS{Reda2022,
  author={Reda, Ameen and Salah, Nada and Adel, John and Ehab, Mohamed and Ahmed, Ismail and Magdy, Mostafa and Khoriba, Ghada and Mohamed, Ensaf Hussein},
  booktitle={2022 2nd International Mobile, Intelligent, and Ubiquitous Computing Conference (MIUCC)}, 
  title={A Hybrid Arabic Text Summarization Approach based on Transformers}, 
  year={2022},
  volume={},
  number={},
  pages={56-62},
  doi={10.1109/MIUCC55081.2022.9781694}}


\end{document}